\documentclass[conference]{IEEEtran}
\IEEEoverridecommandlockouts

\usepackage{cite}
\usepackage{amsmath,amssymb,amsfonts}
\usepackage{algorithmic}
\usepackage{graphicx}
\usepackage{textcomp}
\usepackage{xcolor}
\usepackage{multirow}
\usepackage{threeparttable}
\usepackage{booktabs}
\usepackage{epstopdf}
\usepackage[T1]{fontenc}
\usepackage{aecompl}
\usepackage{cleveref}
\usepackage{tabularx}
\usepackage{dblfloatfix} 
\usepackage{nicematrix,tikz}
\usepackage[linesnumbered, ruled]{algorithm2e}
\SetKwRepeat{Do}{do}{while}%
\usepackage{color, soul}

\def\BibTeX{{\rm B\kern-.05em{\sc i\kern-.025em b}\kern-.08em
    T\kern-.1667em\lower.7ex\hbox{E}\kern-.125emX}}
\begin{document}

\title{Lithium-ion Battery State of Health Estimation by Matrix Profile Empowered Online Knee Onset Identification \\
}

\author{Kate Qi~Zhou,~\IEEEmembership{}
	Yan~Qin,~\IEEEmembership{Member,~IEEE,}
	Chau~Yuen,~\IEEEmembership{Fellow,~IEEE}
	\thanks{This work was supported by EMA-EP011-SLEP-001. (Corresponding author: Yan Qin)}
	\thanks{K. Q. Zhou, Y. Qin are with the Engineering Product Development Pillar, The Singapore University of Technology and Design, 8 Somapah Road, 487372 Singapore. (e-mail: qi$\_$zhou@mymail.sutd.edu.sg,  zdqinyan@gmail.com); C. Yuen is with School of Electrical and Electronics Engineering,  Nanyang Technological University,  50 Nanyang Ave, 639798 Singapore.  (e-mail: chau.yuen@ntu.edu.sg)}}

\onecolumn
\maketitle
\begin{abstract}
Lithium-ion batteries (LiBs) degrade slightly until the knee onset, after which the deterioration accelerates to end of life (EOL).  The knee onset,  which marks the initiation of the accelerated degradation rate, is crucial in providing an early warning of the battery's performance changes.  However, there is only limited literature on online knee onset identification. Furthermore, it is good to perform such identification using easily collected measurements. To solve these challenges, an online knee onset identification method is developed by exploiting the temporal information within the discharge data. First, the temporal dynamics embedded in the discharge voltage cycles from the slight degradation stage are extracted by the dynamic time warping. Second, the anomaly is exposed by Matrix Profile during subsequence similarity search. The knee onset is detected when the temporal dynamics of the new cycle exceed the control limit and the profile index indicates a change in regime. Finally, the identified knee onset is utilized to categorize the battery into long-range or short-range categories by its strong correlation with the battery's EOL cycles. With the support of the battery categorization and the training data acquired under the same statistic distribution, the proposed SOH estimation model achieves enhanced estimation results with a root mean squared error as low as 0.22$\%$.

\end{abstract}


\section{Introduction}
Electric vehicles (EVs) are reshaping the transportation sector, playing a crucial role in paving the path to tackle climate change \cite{52}. Moreover, EVs are also an essential element of intelligent transportation, facilitating transportation safety and mobility. As the principal energy storage for EVs, lithium-ion batteries (LiBs) are a critical component that can be recharged with clean and renewable electricity\cite{57}\cite{33}.  LiB's state of health (SOH) is a measure of the overall condition of the battery,  which is typically expressed by the available capacity as a percentage of the battery's nominal capacity.  Impacted by the internal mechanism and operating environment, LiB's SOH degrades during charging and discharging.  Accurately estimating the battery's SOH is vital and advantageous for continuously monitoring battery life expectancy to avoid EVs' sudden failure, ensuring safety and reliability \cite{21}\cite{56}.

Data-driven way for batteries SOH estimation has emerged as an attractive research field,  as it has the advantage over the model-based way of not knowing the underlying physical condition of batteries \cite{14}\cite{23}. It employs the data collected from historical or real-time measurements and determines the regression relationship by mapping nonlinear functions \cite{50}\cite{20}\cite {27}. There are various approaches reported in the literature for data-driven SOH estimation, such as support vector machine (SVM), autoregressive moving average model (ARMA), feedforward neural network (FNN), convolutional neural network (CNN),  and long-short term memory (LSTM).  For instance,  Greenbank $et\;al. $ \cite{37} proposed a combined feature selection and piecewise-linear modeling method to predict battery capacity degradation. Feng $et\;al. $ \cite{12} used flexible partial charging segment for online SOH estimation based on SVM.  Chen $et\;al. $ \cite{13} achieved accurate SOH prediction by proposing a fusion model based on ARAM and the Elman neural network.  Choi $et\;al. $ \cite{11} utilized multi-channel voltage, current, and temperature data and proposed a capacity estimation framework based on FNN,  CNN, and LSTM. Zhu $et\;al. $ \cite{28} employed XGBoost and used three statistical variables collected from the voltage relaxation curve to estimate the following cycle's capacity.  

However, the varying battery degradation challenges the performance of the aforementioned data-driven methods, in which the training data and testing data are assumed to come from comparable distribution characteristics \cite{32}\cite{22}\cite{48}.  Battery degradation is often observed as a slow decline until it reaches a transition point, namely the knee point, after which it accelerates significantly \cite{7} until it reaches its end-of-life (EOL). The IEEE Standard 485-2020 \cite{10} provides a qualitative description of the knee point as when "the capacity drops slowly throughout the majority of a battery's life,  but starts to degrade quickly in its last phase." 

\begin{figure}[!htb]
\centering
\includegraphics[scale=0.45]{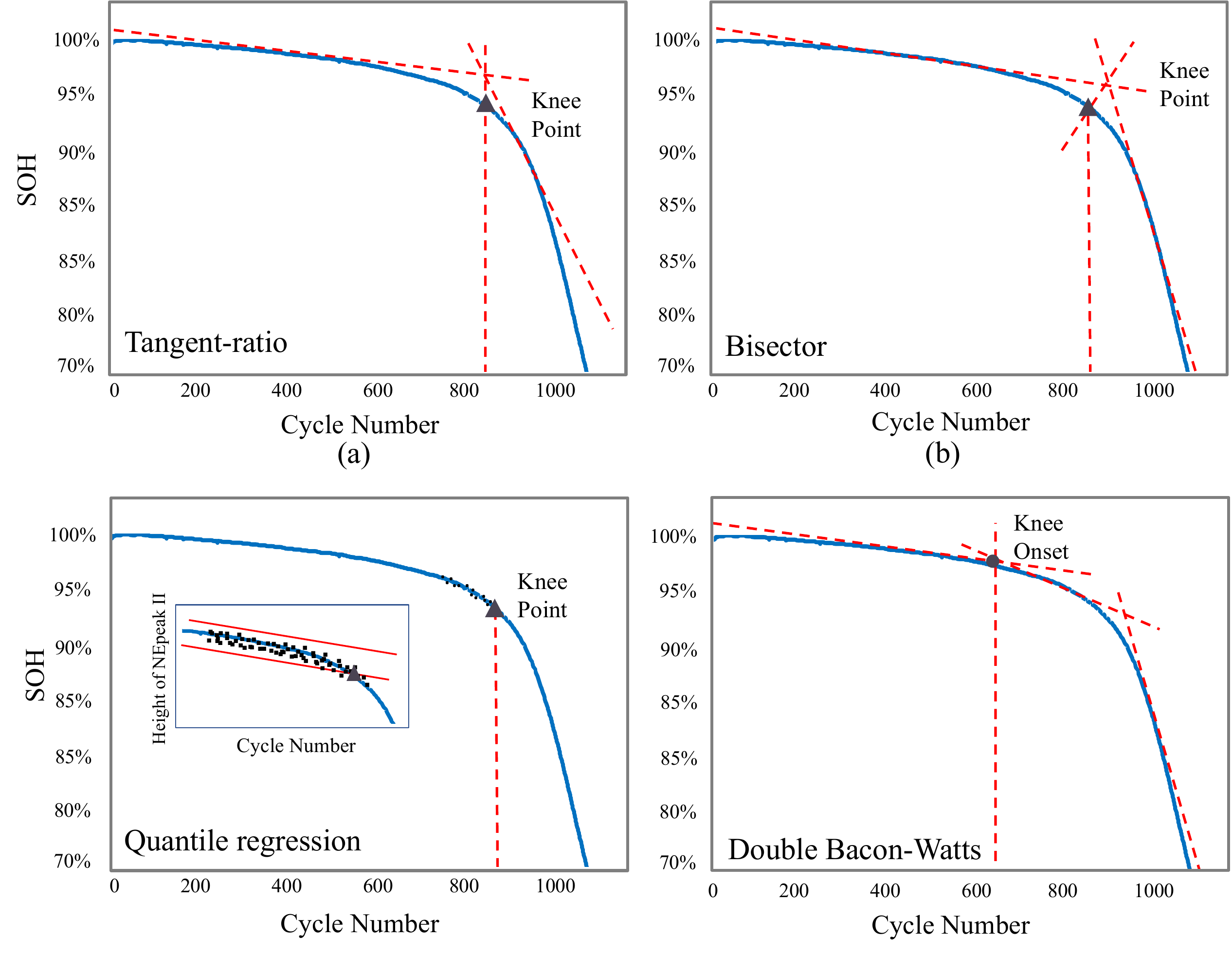}
\caption{Illustration of knee point identification by capacity degradation curve: (a): Tangent-ratio method by Diao $et\;al. $ [22]; (b): Bisector method by Greenbank $et\;al. $ [23]; (c): Illustration of knee point identification using Quantile regression method by Zhang $et\;al. $ [24]; and (d) Illustration of knee onset identification by capacity degradation curve using double Bacon-Watts method by Ferm{\'\i}n-Cueto $et\;al. $ [26].}
\label{MyFigX}
\end{figure}
\addtolength{\textfloatsep}{-0.15in}

Various studies are attempting to identify the knee point throughout the battery's lifespan.  Kim $et\;al. $ \cite{34} proposed a Gradient 1-Knee point curve that can predict the knee point on the capacity curve.  Diao $et\;al. $ \cite{4} calculated the tangent lines based on the minimum and maximum absolute slope-changing ratios and located the knee point on the capacity degradation curve at the intersection of two tangent lines.  Greenbank $et\;al. $ \cite{5} estimated the entire capacity fade trajectory and used the angle bisector of linear regression by fitting the early and late life capacities to find the knee point at the intersection. Zhang $et\;al. $ \cite{6} determined the accelerated fading knee point by combining quantile regression and the Monte Carlo simulation method.  Haris $et\;al. $ \cite{46} detected the knee point in the capacity degradation curve by monitoring the difference between degradation data and a line connected to the initial and current value of the capacity.  Several methods are depicted in Fig. 1(a)-(c) for the purpose of illustration.

 In \cite{2}, a new concept, the "knee onset," was put forward to stand for the point, where it signifies the start of rapid deterioration preceding the knee point.  It is important to note that the knee onset marks the initiation of the accelerated degradation rate preceding the knee point.  It can be conceptualized as the beginning of the knee, with the knee point being the midpoint of this progression. As such, the knee onset is more sensitive to performance degradation in comparison with the traditional knee points.  With the support of battery capacity information, the authors \cite{2} located the knee onset by finding the intersection of two straight lines using the double Bacon and Watts methods, as illustrated in Fig. 1(d).

Although these pioneering researches have achieved remarkable success in identifying the position of the knee point or knee onset, there are still challenges to be further resolved when a reliable online knee onset identification is pursued:

\begin{itemize}

\item The current knee onset identification approaches are conducted offline, which heavily relies on capacity historical data.  The complexities of the internal and external influences makes the lifespans of LiBs different from each other,  yielding different types of knee onset.  Consequently, the offline methods are prohibited from being used for new battery cells when the capacity information is unavailable. 

\item Despite the knee onset providing an early indication of battery health deterioration, online knee onset identification is rarely investigated.  It is beneficial to start monitoring when the battery is initiating a fast degradation phase rather than when it is obviously degraded.

\item The efficacy of the SOH estimation model boosts when the statistical distribution of the training data and testing data are comparable. However, there is limited research on utilizing the location of the knee onset to facilitate SOH estimation and enhance the results.

\end{itemize}

In order to overcome these issues, this work proposes an online knee onset identification method and verifies its efficacy using SOH estimation results.  The temporal dynamics of the discharge voltage cycles are captured by dynamic time warping (DTW), and the abnormality embedded is able to be detected by Matrix Profile (MP) \cite{16},  which signifies the knee onset. The identified knee onset is employed to define the battery life range using the Gaussian mixture model (GMM), which is validated by the correlation analysis. Finally, the knee onset serves as a guide for acquiring the appropriate training data segment and SOH estimation model under the same category,  enabling improved accuracy of SOH estimation.

Our contributions are summarised as follows.

\begin{itemize}
\item A comprehensive investigation of an online method for identifying the knee onset is proposed that utilizes easily obtainable measurements such as raw discharge voltage. This approach is particularly noteworthy as it allows for the identification of knee onset by exploring the internal dynamics of the battery.

\item The proposed online knee onset identification method evaluates the temporal dynamics inherent in the discharging process online,  allowing for the direct detection of system deterioration.  It allows for real-time monitoring of battery performance, providing valuable insights for maintenance and replacement decisions.

\item By leveraging the location of the knee onset,  offline batteries are categorized into different groups as a benchmark for the online battery for SOH estimation model selection.  Furthermore, by selecting training data from the knee onset onward,   the performance of the SOH estimation model is further enhanced by the same statistical distribution data.  This contribution is noteworthy as it allows for accurate and reliable predictions of battery performance in real-world applications.

\end{itemize}

The remainder of this paper is organized as follows.  Section II describes the battery capacity degradation mechanism. Section III illustrates online knee point identification in detail and proposes the SOH estimation model.  Section IV presents the experiment results. The article's conclusion and future work are presented in Section V.

\section{Battery Capacity Degradation with Knee Onset}
The battery cells used in this experiment are manufactured by A123 Systems (APR18650M1A) \cite{19}. They have a nominal capacity of 1.1Ah with a nominal voltage of 3.3V. They are put on a 48-channel Arbin LBT potentiostat in a forced convection temperature chamber of 30$^{\circ}$C. The batteries are charged under a  10-minute fast-charging with one of 224 six-step protocols. The discharge process is done at a discharge current of 4C until the voltage drops from 3.3V to 2V.   39 batteries are chosen from different charging protocols shown in Table I.

\begin{table}[!htb]
	\scriptsize
	\renewcommand{\arraystretch}{1.2}
	\caption{Specification of the Batteries} 
	\vspace{-2mm}
	\label{table_1}
	\begin{center}
\begin{tabular}{p{0.23\textwidth}>{\centering\arraybackslash}m{0.2\textwidth}}
				\toprule  
 				\multicolumn{1}{c} {\textbf {Battery Information}} & \multicolumn{1} {c}{\textbf{Detail}} \\
				\toprule  
				Charge Protocol 1: 5.2-5.2-4.8-4.16C & CH06, CH07 \\
				Charge Protocol 2: 4.4-5.6-5.2-4.252C  & CH08, CH15, CH18, CH32 \\
				Charge Protocol 3: 4.8-5.2-5.2-4.16C & CH01, CH02, CH10, CH20 \\
				Charge Protocol 4: 8.0-4.4-4.4-3.94C & CH13, CH16, CH24, CH47 \\
				Charge Protocol 5: 3.6-6.0-5.6-4.755C & CH11, CH12, CH27,CH29, CH38 \\
				Charge Protocol 6: 6-5.6-4.4-3.834C & CH09, CH21, CH22, CH31, CH26 \\
				Charge Protocol 7: 7-4.8-4.8-3.652C & CH03, CH25, CH26, CH28, CH44 \\
				Charge Protocol 8: 8.0-6.0-4.8-3C & CH14, CH17, CH30, CH25, CH39 \\
				Charge Protocol 9: 8.0-7.0-5.2-2.68C & CH19, CH33, CH34, CH40, CH43 \\
				\toprule
			\end{tabular}
	\end{center}
\vspace{-4mm}
\end{table}

LiB's degradation results from the complex interplay of a variety of chemical and physical components \cite{53}\cite{49} \cite{8} resulting in different capacity degradation curves.  The complete capacity degradation curve of Battery CH32 is shown in Fig. 2(a). The knee onset Cycle $V$ happens when the SOH reduction before the knee onset cycle is less than $\alpha$ and the SOH reduction after is greater than $\alpha$ as follows:
\begin{equation}
\begin{aligned}
& Q_{V-1}-Q_{V-2} < \alpha \\
& Q_{V}-Q_{V-1} \geqslant \alpha \\
\end{aligned}
\end{equation}
where $V$ is the knee onset cycle, $Q_{V}$ is the SOH at the $V^{th}$ cycle.

\begin{figure}[!htb]
\centering
\includegraphics[scale=0.3]{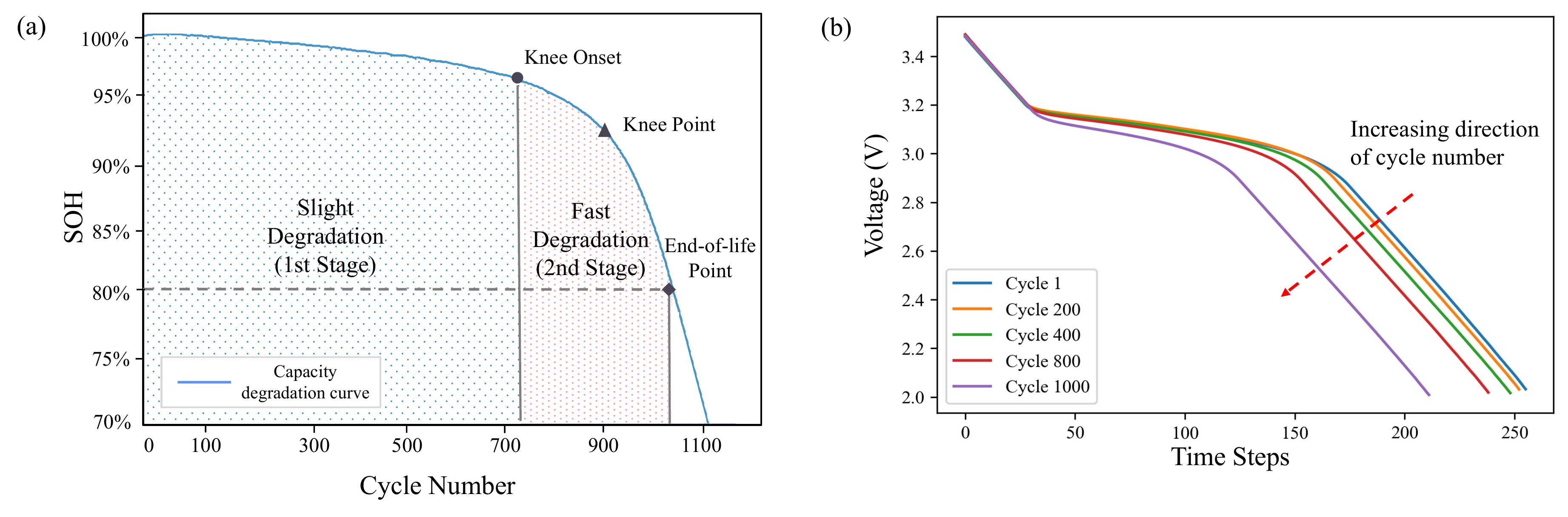}
\caption{(a): The three-stages Lithium-ion battery degradation with the knee onset and end-of-life cycles for CH32 in MIT dataset [28]; (b): Discharge voltage samples in five selected cycles of Battery CH32.}
\label{MyFig2}
\end{figure}

\begin{figure*}[!htb]
\centering
\includegraphics[scale=0.5]{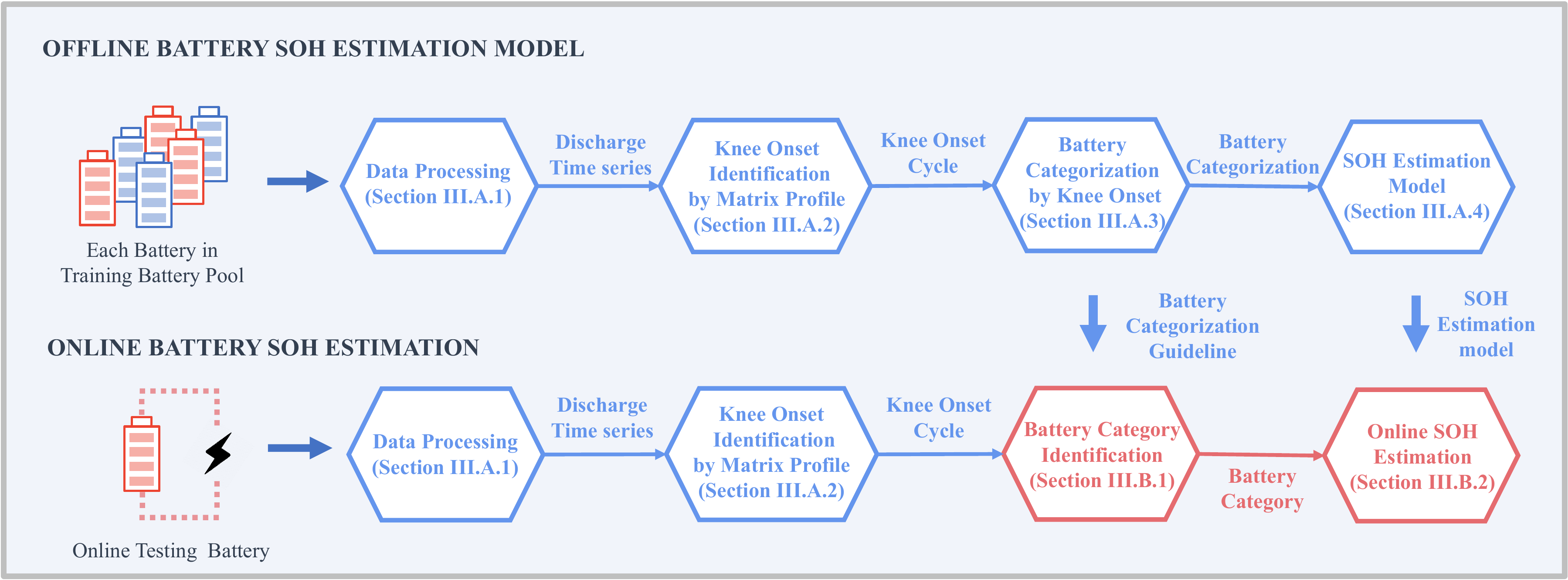}
\caption{Framework of the proposed method}
\label{MyFig3}
\end{figure*}

In online setting, it is not feasible to find the optimal value of $\alpha$ since SOH is not readily available. Therefore, identifying the changes of the internal degradation mechanism within the battery using easily obtainable measurements, like discharge voltage, is worth exploring. As depicted in Fig. 2(b), the discharge voltage curve during battery degradation shows a progressive decrease in time steps and distortion as the battery deteriorates. In this study, the discharge voltage curve is employed to detect the knee onset without the capacity information.

In this paper,  the lifespan of a LiB is divided into three stages based on the knee onset and the EOL cycle.  The $1st$ stage, characterized by a slight decay,  spans from the beginning of the battery's life cycle to the knee onset.   The $2nd$ stage is marked by a rapid deterioration from the knee onset until the EOL. It is critical and requires to be monitored in a real-time manner.  The $3rd$ stage encompasses the post-EOL period.  

\section{Online Battery Knee Onset Identification by Matrix Profile for SOH Estimation}
The process flow, as illustrated in Fig.  3, encompasses both offline and online SOH estimation. Firstly, the knee onset for the group of batteries are identified and categorize the batteries into long-range and short-range.  The respective SOH estimation model is trained using the $2nd$ stage data. Secondly, online knee onset identification is performed for the testing batteries, and the appropriate SOH estimation model is selected based on the category determined by the knee onset position for online SOH estimation.

\subsection{Offline SOH Estimation Model}
\subsubsection{Data Processing}
As batteries degrade, the time steps in the discharging voltage cycles are getting shorter due to charging and discharging processes, resulting in uneven cycle lengths. These uneven cycle lengths are not suitable for data processing and must be synchronized to have the same length while preserving all temporal information \cite{15}.   DTW is a well-established method for identifying the optimal alignment of two time series by assessing their similarity \cite{42}.  It synchronizes two trajectories by expanding and contracting localized segments to achieve a minimum distance between them \cite{41},  constructing a path $\mathbf W$ based on the minimum distance, where $\mathbf W=[w_1, ...,w_g,...,w_G]$ and $w_{g}$ is the $g^{th}$ element of the warping path.

Thus, battery cycle synchronization is achieved by DTW through the following procedure:

Step 1: Reference cycle selection: A reference cycle $\mathbf v_{r}$ is chosen from the battery for the other cycles to synchronize with.  It is denoted as follows:
\begin{equation*}
\begin{aligned}
\mathbf v_{r}=[v_{r}^{(1)},\ldots,v_{r}^{(l)},\ldots,v_{r}^{(d)}]
\end{aligned}
\end{equation*}
where $d$ is the total time steps,  $v_{r}^{(l)}$ is the voltage at time step $g_r^l$ and is represented as $v_{r}^{(l)}(g_r^l,v_r^l)$.

Step 2: DTW path generation: The $k^{th}$ cycle, represented by the vector $\mathbf v_{k}$, is shown as follows:
\begin{equation*}
\begin{aligned}
\mathbf v_{k}=[v_{k}^{(1)},\ldots,v_{k}^{(h)},\ldots,v_{k}^{(t)}]
\end{aligned}
\end{equation*}
where $t$ is the total time steps,  $v_{k}^{(h)}$ is the votlage at time step $g_k^h$ and is represented as $v_{r}^{(l)}(g_k^h,v_k^h)$.

$\mathbf v_{k}$ is synchronized against the reference cycle $\mathbf v_{r}$.  Each time step $v_k^{(h)}(g_{k}^h,v_{k}^h)$ in the cycle searches for the closest value in $\mathbf v_{r}$, such that $v_{k}^h \approx v_{r}^l$.  The time step $g_{k}^h$ for $v_k^{(h)}$ and the time step $g_r^l$ for $v_{r}^{(l)}$ are put into the warping path $\mathbf W$,  where $w_g$ is formed as follows:
\begin{equation*}
\begin{aligned}
w_{g}=(g_{r}^l,g_{k}^h) 
\end{aligned}
\end{equation*}
where $g_{r}^l$ is the time step on the reference cycle and $g_{k}^h$ is the time step on the $k^{th}$ cycle.

Step 3: Cycle synchronization: The warping path is graphically represented by plotting the $g_{r}^l$ values on the x-axis and $g_{k}^h$ values on the y-axis. This allows for the conversion of the voltage $v_k^{(h)}(g_{k}^h,v_{k}^h)$ in the vector $\mathbf v_{k}$ to $x_k^{(h)}(g_{r}^l,g_{k}^h)$.  As a result, the vector $\mathbf v_{k}$ is synchronized to the vector $\mathbf x_{k}$ based on the reference series $\mathbf v_{r}$, where the time step is equal to $d$.  The reference cycle is converted to a diagonal line as $v_r^{(l)}(g_{r}^l,v_{r}^l)$ is converted to $x_r^{(l)}(g_{r}^l,g_{r}^l)$. This results in the vector $\mathbf v_{r}=[v_{r}^{(1)},\ldots,v_{r}^{(l)},\ldots,v_{r}^{(d)}]$ being converted to the vector $\mathbf x_{r}$.  The synchronized cycles are shown as follows and depicted in Fig. 4:

\begin{equation*}
\begin{aligned}
\mathbf x_{k}=[g_{k}^1, \ldots,g_{k}^h,\ldots,g_{k}^t]\\ 
\mathbf x_{r}=[g_{r}^1, \ldots,g_{r}^l,\ldots,g_{r}^d]
\end{aligned}
\end{equation*}
where the total lengths of $\mathbf x_{k}$ and $\mathbf x_{r}$ are $d$.

\begin{figure}[!htb]
\centering
\includegraphics[scale=0.5]{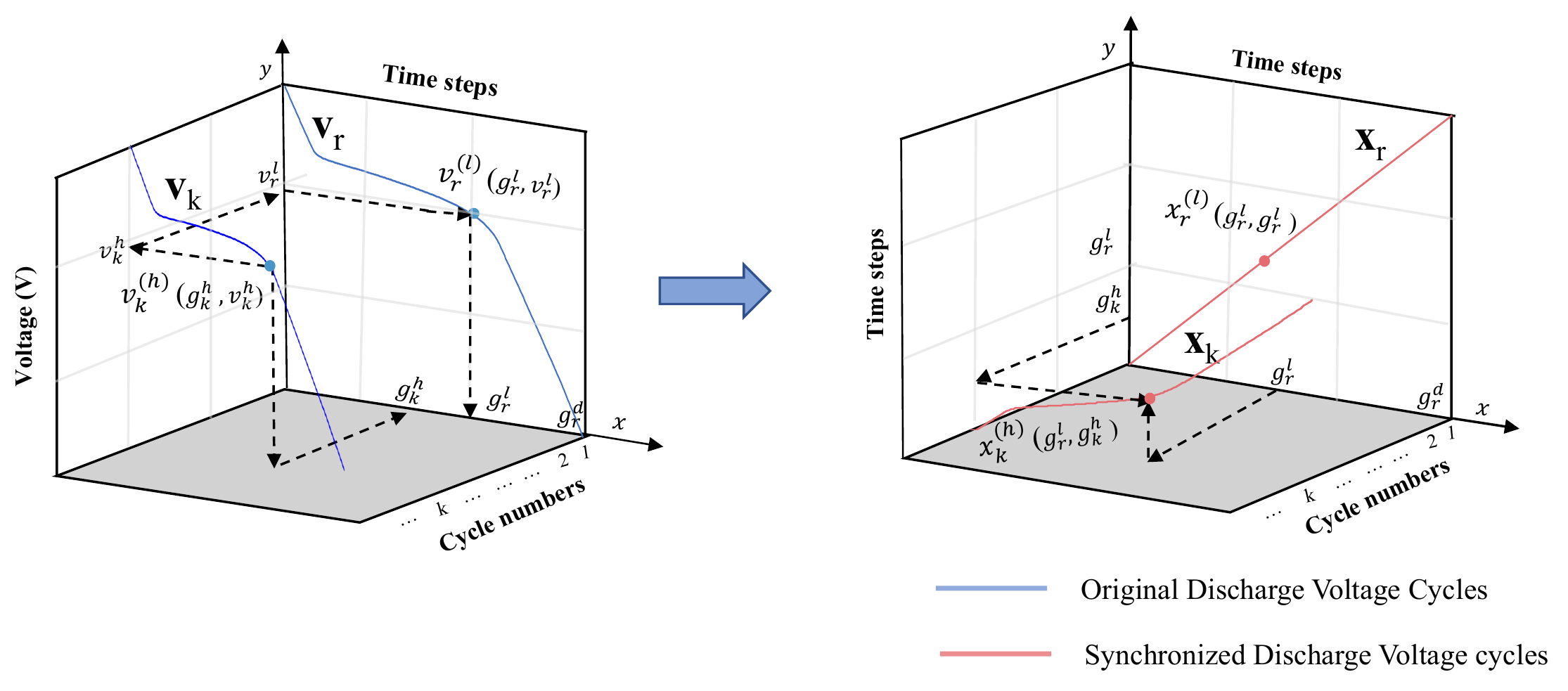}
\caption{Battery discharge cycles synchronized to the same length by DTW}
\label{MyFig4}
\end{figure}

\subsubsection{Knee Onset Identification by Matrix Profile}
The initial $K$ discharge voltage cycles from the slight degradation period are synchronized with the reference cycle following Section III.A.1, resulting in cycles of the same length, $d$.  It is worth that the value of K is much smaller in comparison with the battery's entire life span.  Subsequently,  all $K$ synchronized cycles are concatenated to generate the discharge time series, represented mathematically as $\mathbf T =[\mathbf {x}_{1}, \mathbf {x}_{2} ,\ldots, \mathbf {x}_{K}]$ and denoted as $\mathbf T = \{t_1, t_2,\ldots, t_n\}$, where the length of the time series is calculated as $n=d \times K$.  The discharge time series $\mathbf T$ are analyzed to determine the threshold for detecting incoming cycle anomalies. 

The MP algorithm is a data structure that helps solve the problems of anomaly detection and motif discovery in the time series \cite{18}\cite{43}. It performs this by calculating the Euclidean distance between the subsequences, which is a continuous subset of the time samplings with a user defined query length \cite{47}. The minimum distance is called the nearest neighbor for the subsequence. The nearest neighbors for all the subsequences in the time series are stored into the \textit{matrix profile},  and the locations of the nearest neighbors are recorded in the \textit{profile index}. The motif, which is identified by the lowest point in the \textit{matrix profile}, and the anomaly, which corresponds to the highest point on the \textit{matrix profile}, are illustrated by a synthetic time series in Fig. 5(a).

\begin{figure}[!htb]
\centering
\includegraphics[scale=0.3]{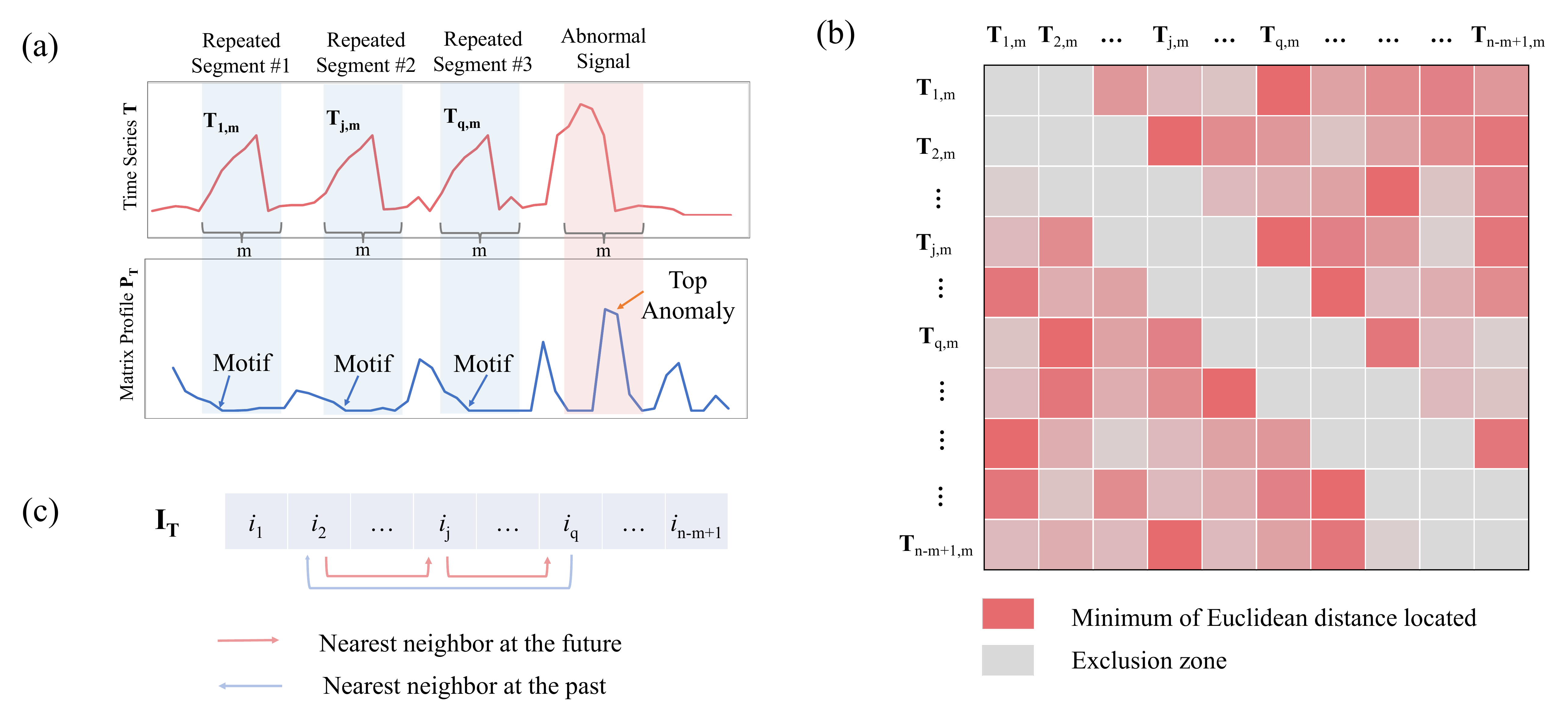}
\caption{Time series Matrix Profile illustraction: (a) \textit{top}: Original time series; \textit{bottom}: \textit{Matrix profile} showing motif and anomaly in the time series; (b) Euclidean distance of the subsequences; (c) \textit{Profile index} shows the location of the nearest neighbour. }
\label{mp}
\end{figure}

In order to generate the \textit{matrix profile} for the battery discharge time series $\mathbf T$, the query length $m$ is determined by choosing the cycle lags $f$ that $m$ equals to $ d \times f$. The subsequence $\mathbf T_{j,m}$ of $\mathbf T$ are consisted of a continuous values starting from position $j$ with length of $m$ that $\mathbf T_{j,m}=\{t_j, t_{j+1}, \ldots, t_{j+m-1}\}$, where $1\leqslant j \leqslant n-m+1$. The Euclidean distances between the $j^{th}$ query subsequence $\mathbf T_{j,m}$ and all other subsequences in the time series $\mathbf T$ are calculated, as illustrated in Fig. 5(b). The minimum distance $p_j$ is calculated as follows:
\begin{equation}
\begin{aligned}
&p_j = \min \limits_{q} \{ |\mathbf T_{j,m} - \mathbf T_{q,m}| \}_{q=1}^{n-m+1} \\
\end{aligned}
\end{equation}

When $q$ equals $j$, it indicates that the subsequence is matching itself, where the distance is calculated from Eq. (2) is zero.  And the subsequence before and after $j$ is closer to zero due to overlapping.  An exclusion zone of $m/2$ is set up to prevent the query from completely or partially matching itself.  As a result, the \textit{matrix profile} is created by storing all the minimum distance values as $\mathbf P_T=\{p_1, p_2,\ldots,p_{n-m+1}\}$.  They are used to annotate the time series $\mathbf T$, with the $j^{th}$ element in the $\mathbf P_T$ representing the nearest neighbor distance for the $j^{th}$ subsequence at $\mathbf T$.

To identify the position of the nearest neighbor,  the index $i_j$ records the location as follows:
\begin{equation}
\begin{aligned}
&i_j = q \\
\end{aligned}
\end{equation}
where $i_j$ is the $j^{th}$ index and $q$ is obtained from the subsequence $\mathbf T_{q,m}$,  who is the nearest neighbor of $\mathbf T_{j,m}$.

The \textit{profile index} is formed as $\mathbf I_T=\{i_1,i_2,\ldots,  i_{n-m+1}\}$ to record the index information for each location that the nearest neighbor distance is located,  whether it is in the past or in the future, as shown in Fig.  5(c).

In order to extract the cycle-based \textit{matrix profile} and \textit{profile index} information,  every $d^{th}$ value from $\mathbf P_T$ and $\mathbf I_T$ is taken and form cycle-based $\mathbf P_C=\{p_d,  p_{2d}, \ldots, p_{(K-f)\times d}\}$ and cycle-based $\mathbf I_C=\{i_d, i_{2d}, \ldots,i_{(K-f)\times d}\}$ and they are denoted as follows:

\begin{equation}
\begin{aligned}
&\mathbf P_C=\{\tilde{p}_1,  \tilde{p}_2, \ldots, \tilde{p}_{K-f}\}\\
&\mathbf I_C=\{\tilde{i}_1, \tilde{i}_2, \ldots,\tilde{i}_{K-f}\}
\end{aligned}
\end{equation}

When the incoming discharge cycle acts differently from prior cycles, $p_j$ increases. It reduces again once it identifies a subsequent cycle with similar behavior. When the battery deterioration transits from the $1st$ stage to the $2nd$ stage, $p_j$ increases quickly as the new cycle enters a fast degradation period in which the similarity between the two stages is low,  as reflected in the increasing distance of the nearest neighbor.  Thus,  an upper control limit (UCL) is established from the slight degradation stage.  Once the value $p_j$ exceeds the UCL,  an anomaly is considered to be detected.  As $\mathbf P_C$ is generated by the $K$ cycles from the $1st$ stage, during which the battery slightly degrades, the UCL is calculated as follows:

\begin{equation}
\begin{aligned}
UCL= \overline {\tilde{p}_{j}}|_{j=2}^{K-f} +1.5\sigma
\end{aligned}
\end{equation}
where $\overline {\tilde{p}_{j}}|_{j=2}^{K-f}$ is the mean and $\sigma$ is the standard variance of the \textit{matrix profile} $\mathbf P_C$.

Other than the \textit{matrix profile} exceeding UCL, to be a transit point, the nearest neighbor of the knee onset cycle and the cycles after should be pointing to the future as the battery degradation is entering the $2nd$ stage.  Furthermore, the nearest neighbors of the cycles before knee onset should be pointing to the past as they belong to $1st$ stage. 

Thus,  the knee onset identification is proceeded as follows:

\begin{itemize}

\item Setp 1: Initiate $s=K$.  The cycles after $K$ cycles are monitored.   Let $s=s+1$. $\mathbf v_{s}$ is converted into synchronized cycle $\mathbf x_{s}$ and concatenated to $\mathbf T$.   As the  \textit{matrix profile} has a delay of $f$ cycles,  discharge voltage cycle $s$ is collected for analyzing the $s-f$ cycle.  

\item Step 2: $\mathbf P_C$ and $\mathbf I_C$ following Eqs. (2)-(4) are generated and $\tilde{p}_{s-f}$  are compared with the UCL.  

\item Step 3: If $\tilde{p}_{s-f}$ exceeds UCL,  the \textit{profile index} switching point is searched.  If the $\tilde{i}_{s-f}$ are pointing to the future cycle and $\tilde{i}_{s-f-1}$ is pointing to the past or $\tilde{i}_{s-f}$ are pointing to the past cycle and the next cycle $\tilde{i}_{s-f+1}$ is pointing to the future,  the next $f$ cycles are being monitored.  If all the $f$ cycles are pointing to the cycles after $s-f-1$,  it is considered that the discharge voltage cycles have entered a new regime that the knee onset happened at $s-f$.  The knee onset cycle is denoted as $\vartheta$. The details of the knee onset identification are described in Algorithm 1.  
\end{itemize}

\begin{algorithm}
\begin{algorithmic}[1]
\STATE Initialize $s=K$;\\
  \STATE s=s+1;\\
   \STATE Input: $\mathbf v_{s}$;\\
   \STATE Convert $\mathbf v_{s}$ to $\mathbf x_{s}$ to concatenate to $\mathbf T$;\\
   \STATE Generate $\mathbf P_C$ and $\mathbf I_C$ according to Eqs. (2)-(4);\\

\IF{\textit{matrix profile} $\tilde{p}_{s-f} $ < UCL}
\STATE Go to Step 2
\ELSE 
\STATE Check \textit{profile index}
\IF {$ \tilde{i}_{s-f-1} <(s-f-1)$  $\mathbf {and}$ $\tilde{i}_{s-f} \geqslant  s-f$ $\mathbf {or}$  \\
$ \tilde{i}_{s-f} <(s-f)$  $\mathbf {and}$ $\tilde{i}_{s-f+1} \geqslant  s-f+1$ }
\STATE Check the next $f$ cycles
\WHILE{$e=1,2,3,..., f$}
\STATE $u=s+e$
   \STATE Convert $\mathbf v_{u}$ to $\mathbf x_{u}$ to concatenate to $\mathbf T$;\\
   \STATE Generate $\mathbf P_C$ and $\mathbf I_C$ according to Eqs. (2)-(4);\\
\IF{$\tilde{i}_{u-f} < (s-f-1)$ }
\STATE Go to Step 2
\ENDIF
\ENDWHILE
\ELSE 
\STATE $s-f$ is identified as the knee onset cycle $\vartheta$
\ENDIF
\ENDIF
  \caption{LiB Knee Onset Identification}
\end{algorithmic}
\end{algorithm}

\begin{figure}[!htb]
\centering
\includegraphics[scale=0.6]{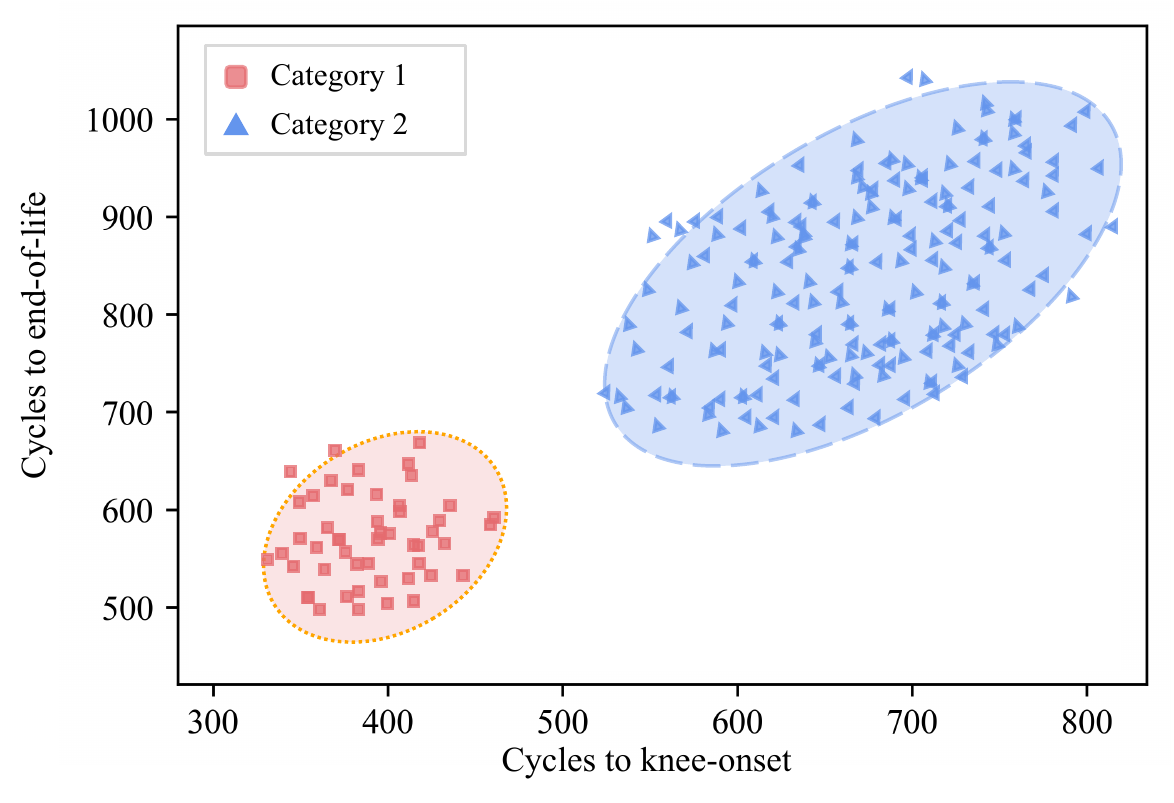}
\caption{Illustration of battery clustering using synthetic data by Gaussian mixture model}
\label{cluster}
\end{figure}

\subsubsection{Battery Categorization by Knee Onset}

The linear relationship between the knee onset cycle $\vartheta$ and the EOL cycle $\varepsilon$ is investigated by the coefficient of determination R-Squared (R$^2$).  A linear regression line is applied to the offline batteries in the pool,  where the estimated EOL cycle $\hat{\varepsilon_\gamma}$ of $\gamma^{th}$ battery is obtained by the linear function $f(\cdot)$ that $\hat{\varepsilon_\gamma}=f(\vartheta_\gamma,\beta)+e_\gamma$.

The coefficient of determination R-Squared (R$^2$) is calculated as follows:

\begin{equation}
\begin{aligned}
R_{pool}^2 = 1-\frac{\sum \left ( \varepsilon_{\gamma}-\hat{\varepsilon}_\gamma \right )^2}{\sum \left ( \varepsilon_\gamma-\overline{\varepsilon} \right )^2}
\end{aligned}
\end{equation}
where $\varepsilon_\gamma$ is actual EOL cycle and $\hat{\varepsilon}_\gamma$ is the estimated EOL cycle of the $\gamma^{th}$ battery in the pool,   and $\overline{\varepsilon}$ is the average of the actual EOL cycles of all the batteries in the pool.

\begin{figure}[!htb]
\centering
\includegraphics[scale=0.52]{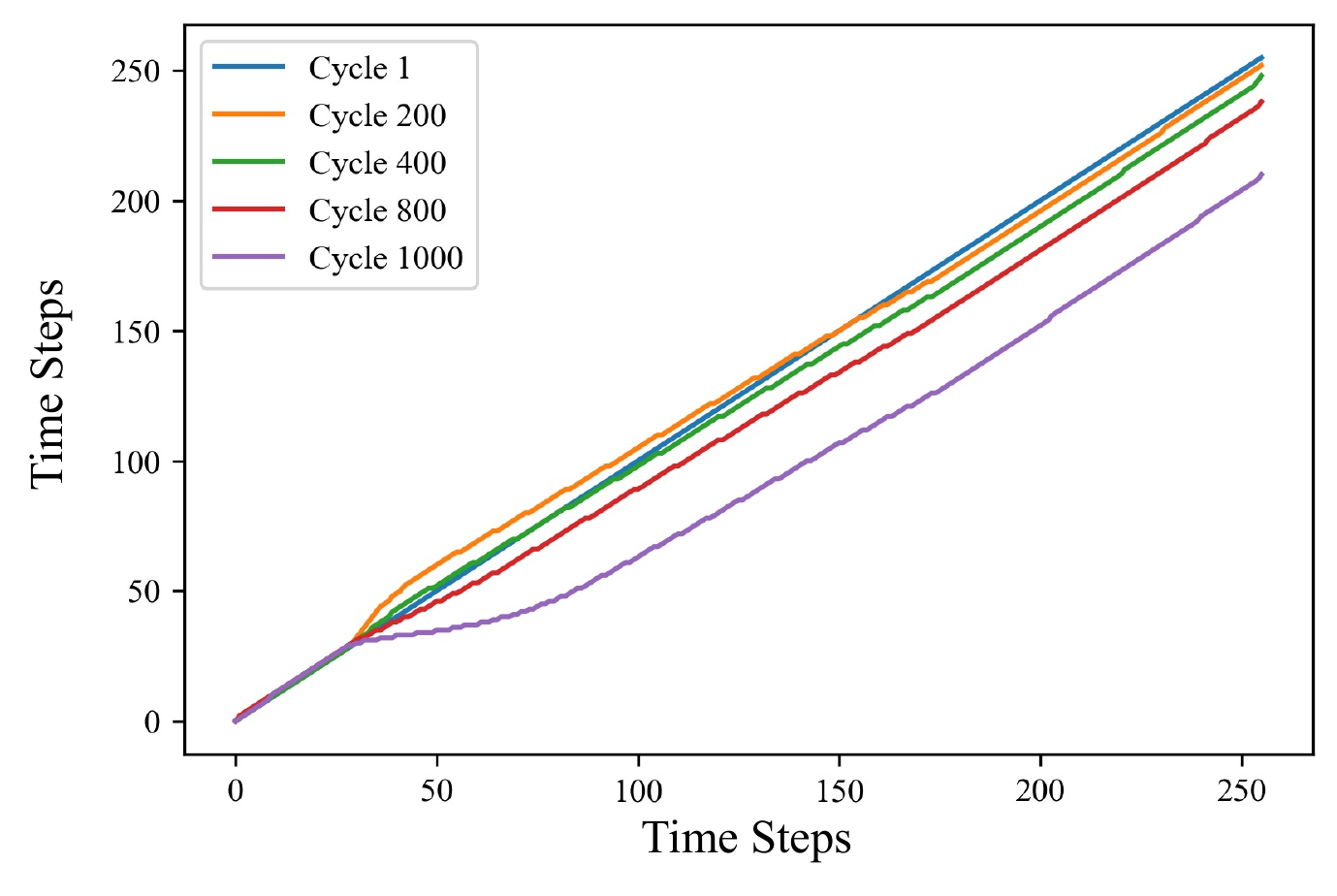}
\caption{Sample of discharge voltage cycle synchronization by DTW of Battery CH32}
\label{dtw}
\end{figure}

The value of $R_{pool}^2$ ranges from 0 to 1. When $R_{pool}^2$ is above 90$\%$,  implying more than 90$\%$ of the dependent variables are explainable by the independent variables \cite{44}, indicating a strong linear correlation between knee onset and EOL cycles.  Therefore, the knee onset may serve as an early indicator for the battery's EOL.  

The distribution of the knee onset and  EOL cycles is utilized to classify the offline batteries into different categories by data clustering.  Data clustering is the process of identifying and differentiating the characteristics of distinct groups within a data collection \cite{26}. The GMM is an unsupervised learning method to determine the data clustering, where its probability density function (PDF) is the mixture of the weighted sum of densities in the group \cite{36}. 

For a given a dataset $A:\{a_1, a_2, ..., a_{\Gamma} \} $ and $B$ predefined clusters, the PDF is as follows: 
\begin{equation}
\begin{aligned}
f(a)=\sum_{b=1}^{B} \pi_{b} p(a|\mu_{b} ,\delta_{b})\\
\end{aligned}
\end{equation}
where $\pi_{b}(b=1, ..., B) $ are the weights that $\sum_{b=1}^{B} \pi_{b}=1$, $\mu_{b}$ is the mean and $\delta_{b}$ is the covariance for each cluster. 

GMM utilizes the expectation-maximization algorithm to fit the mixture-of-Gaussian models to find the parameters $\theta_{b} = (\pi_{b}, \mu_{b}, \delta_{b})$ in order to maximize the log likelihood as follows \cite{40}:
\begin{equation}
\begin{aligned}
logL(\theta, a)= \sum_{\zeta=1}^{\Gamma} log \sum_{b=1}^{B} \pi_{b} f(a_{\zeta}|\mu_{b} ,\delta_{b})\\
\end{aligned}
\end{equation}

Fig.  6 illustrates the clustering outcome from synthetic data,  which separates the offline batteries in the pool into two categories regarding lifespan. 

\subsubsection{SOH Estimation Model}
In this section, a SOH estimation model is proposed, which incorporates the concept of knee onset as a crucial reference point.

The model is developed by first collecting and synchronized the initial $K$ discharge cycles of the training, resulting in $\mathbf T^{tr} =[\mathbf {x}^{tr}_ {1}, \mathbf {x}^{tr}_{2} ,\ldots, \mathbf {x}^{tr}_{K}]$, as detailed in Section III.A.1.  Subsequently, the $UCL^{tr}$ of the matrix profile is established through the methodology detailed in Section III.A.2. The knee onset cycle $\vartheta^{tr}$ is identified for each offline battery.

\begin{figure*}[!htb]
\centering
\includegraphics[scale=0.45]{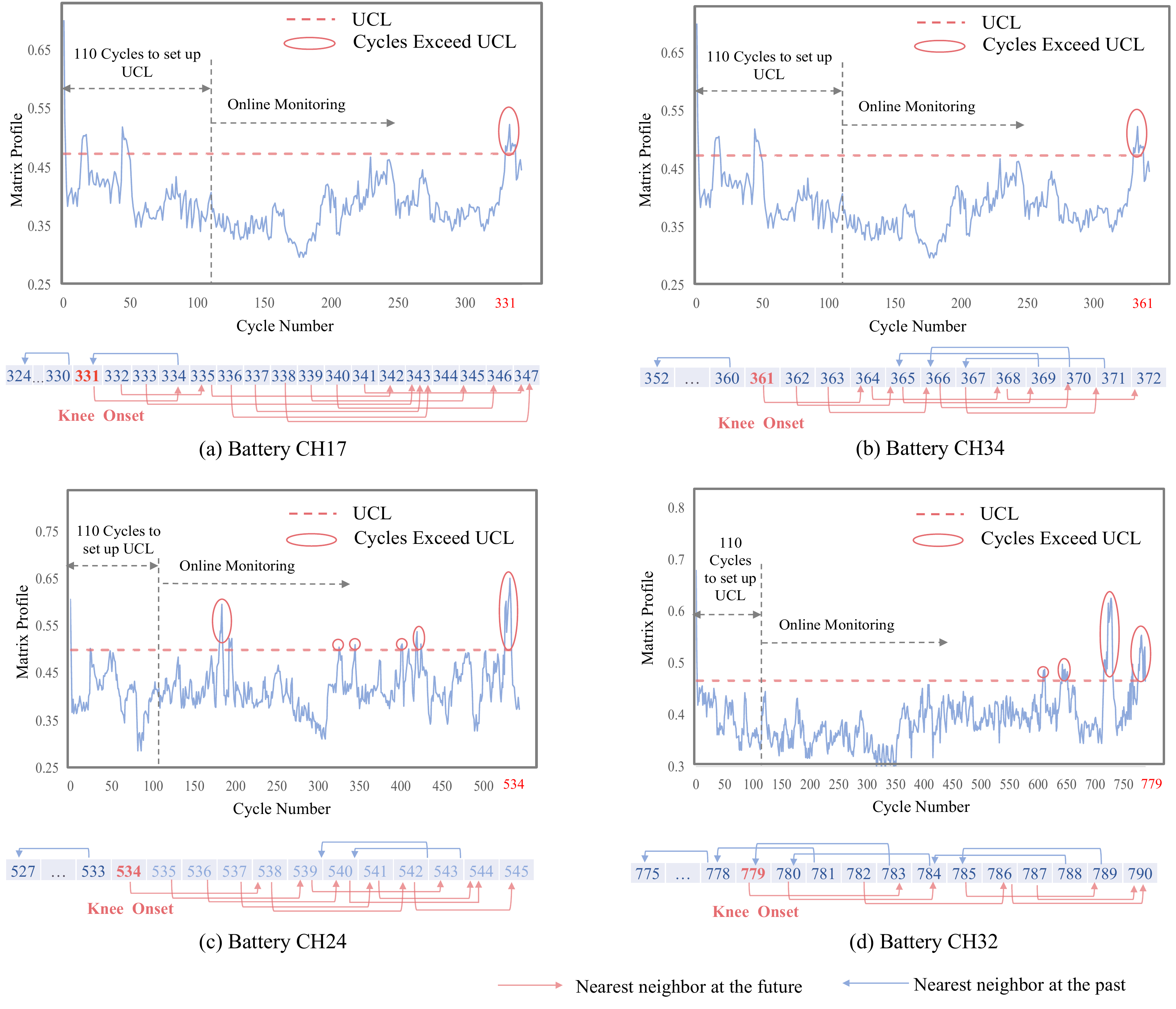}
\caption{(a) Battery CH17, Top:  \textit{Matrix profile} and UCL, Bottom: Knee onset identification by \textit{profile index}; (b) Battery CH34, Top:   \textit{Matrix profile} and UCL, Bottom: Knee onset identification by \textit{profile index}; (c) Battery CH24, Top:   \textit{Matrix profile} and UCL, Bottom: Knee onset identification by \textit{profile index}; (d) Battery CH32, Top:  \ \textit{Matrix profile} and UCL, Bottom: Knee onset identification by \textit{profile index}. }
\label{MyFig8}
\end{figure*}

Leverage the knee onset cycle $\vartheta^{tr}$, the offline batteries are categorized into $B$ clusters, with one battery is chosen from each cluster to train the offline SOH estimation model for each category.  The training data was limited to the data from the knee onset cycle $\vartheta^{tr}$ to the EOL cycle $\varepsilon^{tr}$ only. 

LSTM has been implemented in battery SOH estimation in prevailing research and is achieving good accuracy \cite{24} \cite{25}.  In this work, LSTM is utilized to establish the relationship between the synchronized cycle $\mathbf x^{tr}_s$ and the SOH. The proposed model is a multiplayer LSTM consisting of an input layer, several LSTM layers, and a dense layer before the output layer to minimize the estimation error as shown in Eqs. (9)-(10). 

\begin{equation}
\begin{aligned}
&\min\sum{(Y^{tr}_s-\hat{Y}^{tr}_s)}^2\\
\end{aligned}
\end{equation}
\begin{equation}
\begin{aligned}
&\mathbf {s.t.  } \hat{Y}^{tr}_s= F^b_{LSTM}(\mathbf x^{tr}_s)
\end{aligned}
\end{equation}
where $\mathbf x^{tr}_s$ is the synchronized cycle from discharge cycle $\mathbf v^{tr}_s$, $\hat{Y}^{tr}_s$ is the estimated capacity, and $Y^{tr}_s$ is the measured capacity of the offline training battery, $s \in [\vartheta^{tr},\varepsilon^{tr}]$ ; $F^b_{LSTM}(\cdot)$ stands for the offline LSTM model for Cluster $b$, $b \in [1,B]$

\subsection{Online Battery Category Identification and SOH Estimation}
\subsubsection{Battery Category Identification}
The procedure for online knee onset identification within the testing battery is executed as follows: First, the initial $K$ discharge cycles of the testing battery are collected and synchronized, resulting in $\mathbf T^{te} =[\mathbf {x}^{te}_ {1}, \mathbf {x}^{te}_{2} ,\ldots, \mathbf {x}^{te}_{K}]$,  as detailed in Section III.A.1. Subsequently, the $UCL^{te}$ of the matrix profile is established through the method outlined in Section III.A.2.  The synchronized cycles $\mathbf x^{te}_s$ are monitored beyond the initial $K$ cycles, and the online knee onset $\vartheta^{te}$ is detected when the \textit{matrix profile} of the current discharge cycle exceeds its $UCL^{te}$ and the \textit{profile index} indicates a nearest neighbor no longer originating from the past.

Based on the cluster analysis presented in Section III.A.3, the category of the testing battery is determined by the position of the knee onset cycle $\vartheta^{te}$. 

\subsubsection{Online SOH Estimation}
The appropriate SOH estimation model is selected according to the category the online battery is in.  The discharge voltage cycle after the knee onset cycle $\vartheta^{te}$ is synchronized to $\mathbf x^{te}_s$ and fed into the chosen model, namely the $F^b_{LSTM}(\cdot)$, to generate the online SOH estimation as follows:

\begin{equation}
\begin{aligned}
&  \hat{Y}^{te}_s= F^b_{LSTM}(\mathbf x^{te}_s)
\end{aligned}
\end{equation}
where $\mathbf x^{te}_s$ is the $s^{th}$ synchronized discharge voltage cycle and $\hat{Y}^{te}_s$ is the estimated SOH of the $s^{th}$ cycle of online battery, $s \in [\vartheta^{te},\varepsilon^{te}]$.

\section{Experiment Results and Discussion}
The efficacy of the proposed method is presented in this section. This paper uses 39 batteries from the MIT dataset \cite{19} to verify our proposed method.  Among them, 4 batteries from different charging protocols, namely CH17, CH34, CH24, and CH32, are randomly chosen and used to demonstrate the process of finding the knee onset as well as establish the offline SOH estimation model.

\subsection{Offline SOH Estimation Model}
\subsubsection{Data Processing} Following the procedures outlined in Section II.A.1, the first discharge voltage cycle from CH17, CH24, CH32, and CH34, respectively, is chosen as the reference cycle for the rest of the cycles to be transformed into synchronized cycles by DTW. The discharge voltage cycles are synchronized to have the same length as the reference cycles: $d_{CH17}=260$,  $d_{CH34}=258$, $d_{CH24}=260 $, and $d_{CH32}=256$.  Result for $d_{CH32}=256$ is shown in Fig. 7 for illustration, compared with the original discharge cycle shown in Fig. 2(b).

\subsubsection{Knee Onset Identification by Matrix Profile}
The first 110 synchronized cycles from each battery under slight degradation stage are used to generate the \textit{matrix profile} UCL following Eqs. (2) to (5). Cycle lap $f$=10 is chosen. The UCLs for Batteries CH17, CH34, CH24, and CH32 are 0.474, 0.487, 0.476,  and 0.456,  respectively.

After that, the cycles after the 110 cycles are collected and follow Algorithm 1 to perform knee onset identification. The cycle based \textit{matrix profile} $\mathbf {P_C}$ of each cycle is compared with the UCL until the cycle exceeds the UCL. At the same time, the \textit{profile index} is checked for the next $10$ cycles to detect regime changes. Details of the \textit{matrix profile} and \textit{profile index} checking is shown in Fig. 8. Using Battery CH17 as an example, a few \textit{matrix profiles} $\tilde{p}$ surpass the UCL of 0.474, but the \textit{profile index} $\tilde{i}$ is not meeting the criteria of index changes until Cycle 331. $\tilde{i}_{331}$ is 334 indicates that Cycle 331 is more similar to future cycle and $\tilde{i}_{330}$ is 324 indicates that Cycle 330 is more similar to the past cycle. It indicates a shift in pattern from Cycle 330 to Cycle 331. And the \textit{profile index} of 10 cycles after Cycle 331 all points to the cycles after Cycle 330. It infers that all the cycles after Cycle 331 are entering a new stage where they are no longer similar to the past. Thus, knee onset is at Cycle 331. Similarly, the knee onsets for Batteries CH34, CH24, and CH32 are identified as Cycles 361, 534, and 779, respectively.

As shown in Table II, different cycle lags $f$ of 6, 20, and 30 cycles are chosen to verify the location of the knee onset. The maximum gap of knee onset cycles detected by different cycle lags is 18 cycles, indicating that the proposed method is practical.

\begin{table}[]
\begin{center}
\caption{knee Onset Cycle by Different Cycle Lags}
\begin{tabular}{ccccc}
\toprule
Cycle Lags     & \multicolumn{1}{c}{CH17} & \multicolumn{1}{c}{CH34} & \multicolumn{1}{c}{CH24} & CH32 \\ \hline
\\[-1em]
$f=6$             & \multicolumn{1}{c}{336}  & \multicolumn{1}{c}{366}  & \multicolumn{1}{c}{535}  & 781  \\ \hline
\\[-1em]
$f=10$            & \multicolumn{1}{c}{331}  & \multicolumn{1}{c}{361}  & \multicolumn{1}{c}{534}  & 779  \\ \hline
\\[-1em]
$f=20$            & \multicolumn{1}{c}{349}  & \multicolumn{1}{c}{359}  & \multicolumn{1}{c}{531}  & 774  \\ \hline
\\[-1em]
Max Cycle Gap & \multicolumn{1}{c}{18}    & \multicolumn{1}{c}{7}    & \multicolumn{1}{c}{4}    & 7    \\ 
\toprule
\end{tabular}
\end{center}
\end{table}

\begin{figure}[!htb]
\centering
\includegraphics[scale=0.6]{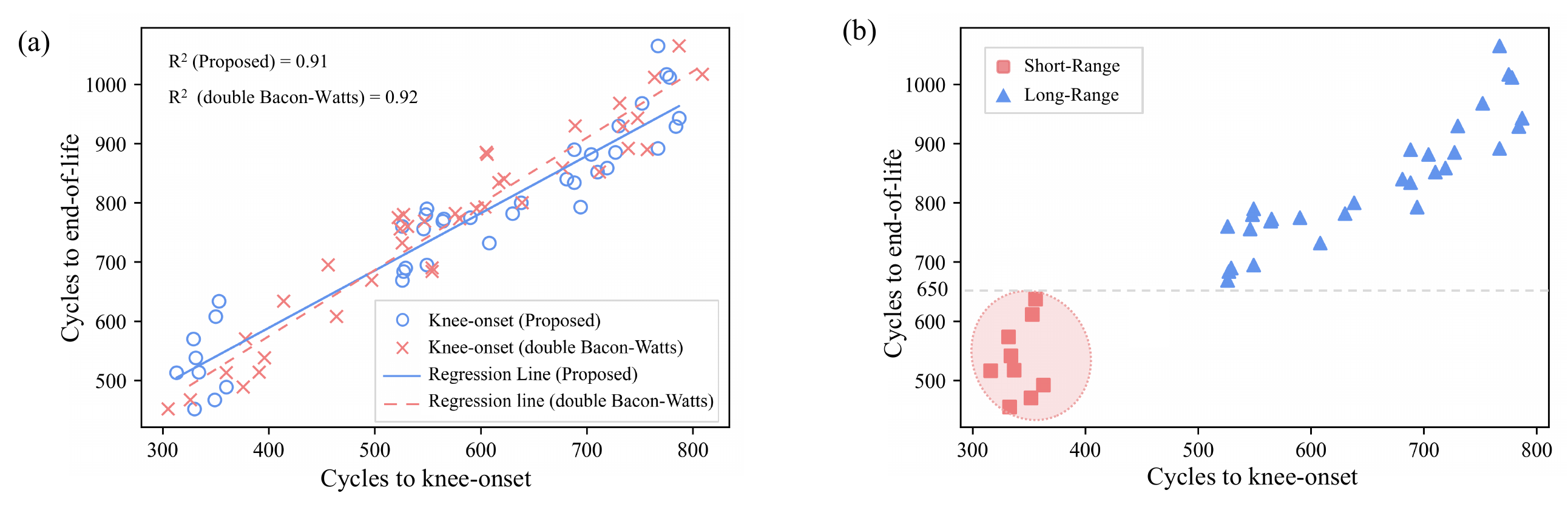}
\caption{Relationship mapping between knee onset and EOL cycles for 39 batteries (a) $R^2$ comparison between proposed methods and double Bacon-Watts methods, (b) Battery categorization by Gaussian mixture model based on knee onset and EOL cycles distribution}
\label{R2}
\end{figure}

\subsubsection{Battery Categorization by Knee Onset}
The knee onsets are identified for 39 batteries from the pool.  A linear regression line is plotted based on their knee onset cycles and EOL cycles, as shown in Fig. 9(a). To benchmark the proposed methods, the knee onsets identified by the double Bacon-Watts method \cite{2}, where the knee onsets are identified by the capacity curve, are plotted. The $R^2$ calculated based on Eq. (6) from the double Bacon-Watts methods is 0.92,  and the $R^2$ from the proposed methods is 0.91. It suggests the knee onset identified by the proposed online identification methods achieves a good linear relationship between the knee onset cycle and the EOL cycle, which is comparable to the offline method that is based on the capacity curve.

\begin{table*}[]
\scriptsize
\renewcommand{\arraystretch}{1.2}
\caption{Performance comparison regarding RMSE between the proposed method and its counterparts. }
\vspace{-3mm}
\label{table_4}
\begin{center}
\setlength{\tabcolsep}{4.4pt}
\begin{tabular}{>{\centering\arraybackslash}m{0.15\textwidth}|>{\centering\arraybackslash}m{0.05\textwidth}>{\centering\arraybackslash}m{0.05\textwidth}>{\centering\arraybackslash}m{0.05\textwidth}>{\centering\arraybackslash}m{0.05\textwidth}|>{\centering\arraybackslash}m{0.05\textwidth}>{\centering\arraybackslash}m{0.05\textwidth}>{\centering\arraybackslash}m{0.05\textwidth}>{\centering\arraybackslash}m{0.05\textwidth}|>{\centering\arraybackslash}m{0.05\textwidth}>{\centering\arraybackslash}m{0.05\textwidth}>{\centering\arraybackslash}m{0.05\textwidth}>{\centering\arraybackslash}m{0.05\textwidth}}
\toprule
                         & \multicolumn{4}{c|}{The Proposed   Method} & \multicolumn{4}{c|}{Comparison 1} & \multicolumn{4}{c}{Comparison 2} \\ \toprule
 Offline Battery           & CH17& CH34& CH24& CH32& CH17& CH34& CH24& CH32& CH17 & CH34& CH24& CH32 \\ \toprule
Offline  Battery   Category & Short    & Short    & Long    & Long   & Short           & Short           & Long           & Long           & Short         & Short         & Long         & Long         \\ \hline
Online Battery   Category & Short    & Short    & Long    & Long   & Long            & Long            & Short          & Short          & Short         & Short         & Long         & Long         \\ \toprule
Training Data  Range    & \multicolumn{4}{c|}{$2nd$ Stage}          & \multicolumn{4}{c|}{$2nd$ Stage}                                       & \multicolumn{4}{c}{$1st$ and $2nd$ Stage}                       \\ \hline
RMSE (Minimum)                  &     $\mathbf {0.0036 } $ &   $\mathbf {0.0039  } $     &     $\mathbf {0.0029 } $   &  $\mathbf {0.0022  } $    &    0.0054             &       0.0100          &     0.0090           &    0.0154            &  0.0055             &    0.0059           &     0.0048         &      0.0024        \\ 
RMSE (Median)               &       $\mathbf {0.0059 } $  &   $\mathbf { 0.0072  } $    &    $\mathbf {0.0082 } $    &  $\mathbf { 0.0054   } $  &         0.0230        &    0.0289             &      0.0218          &       0.0272         &   0.0107           &   0.0115            &      0.0187        &     0.0078         \\ 
RMSE (Maximum)                &   $\mathbf { 0.0151 } $    &     $\mathbf { 0.0145 } $   &   $\mathbf {  0.0220 } $   &  $\mathbf {  0.0181} $    &          0.0394       &       0.0445          &      0.0320          &     0.0361           & 0.0185              &     0.0227          &      0.0348        &    0.0200          \\ 
\toprule
\end{tabular}
\begin{tablenotes}\scriptsize
\item[] Comparison 1: Training and Testing Batteries in Different Categories\\
Comparison 2: Training  Data Using the Entire Cycle Life
\end{tablenotes}
\end{center}
\end{table*}

Furthermore, two clusters are observed in Fig. 9(b). The first cluster contains batteries with EOL cycle greater than 650 cycles.  And the second cluster is composed of batteries with EOL cycle smaller than 650 cycles.  Here, the batteries in the first category are referred to as long-range, and those distributed in the second category are considered short-range.  Based on that, CH17 and CH34 are categorized as short-range batteries, while CH24 and CH32 are grouped as long-range batteries.

\subsubsection{SOH Estimation Model}
Once the knee onset of the battery has been determined, the training data from the knee onset through EOL cycles, referred to as $2nd$ stage,  is gathered to train the SOH estimation model. The SOH estimation model is trained for Battery CH17, CH34, CH24, and CH32 separately, where the models from Battery CH17 and CH34 are used for short-range batteries and the ones from Battery CH24 and CH32 are used for long-range batteries. The LSTM-based model is comprised of two LSTM layers of 300 and 500 neurons. A normal layer of 100 neurons is added before the output layer. 

\begin{figure}[!htb]
\centering
\includegraphics[scale=0.38]{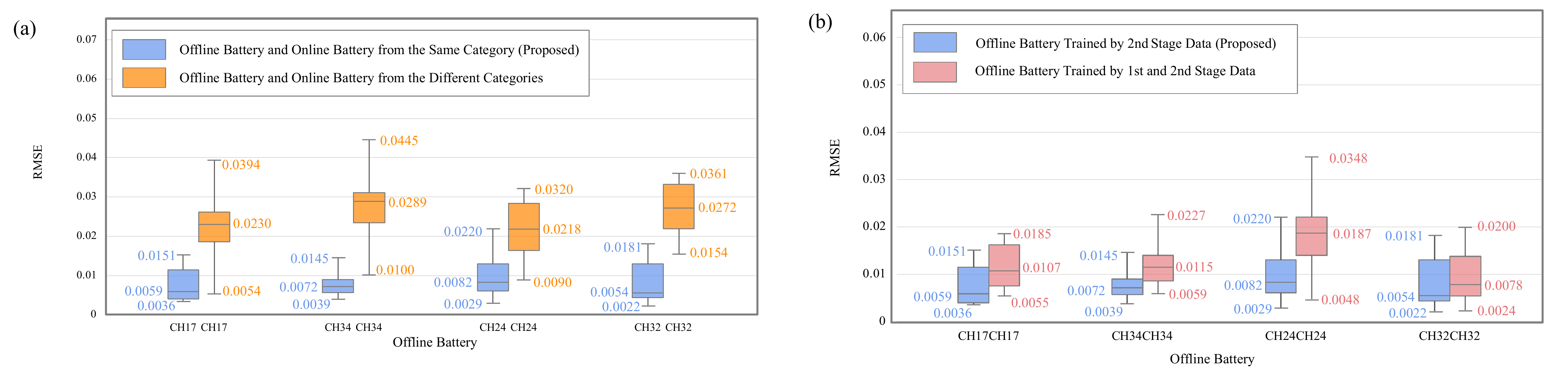}
\caption{Box-plots showing the distribution error: (a) $RMSEs$ between the batteries from the same category and the batteries from different categories; (b) $RMSEs$ between the proposed methods and models trained by the entire life cycle.}
\label{rmse}
\end{figure}

The model is trained by minimizing the root mean squared error (RMSE) as follows:
\begin{equation}
\begin{aligned}
\mathbf {RMSE}=\sqrt{\frac{1}{\varepsilon-\vartheta}\sum_{s=\vartheta}^{\varepsilon}{(Y^{tr}_s -\hat{Y}^{tr}_s)^2}}
\end{aligned}
\end{equation}
where $\vartheta$ is the knee onset, $ \varepsilon$ is the EOL cycle,  $\hat{Y}^{tr}_s$ is the estimated SOH,  and $Y^{tr}s$ is the measured SOH of the battery.

\begin{figure*}[!htb]
\centering
\includegraphics[scale=0.45]{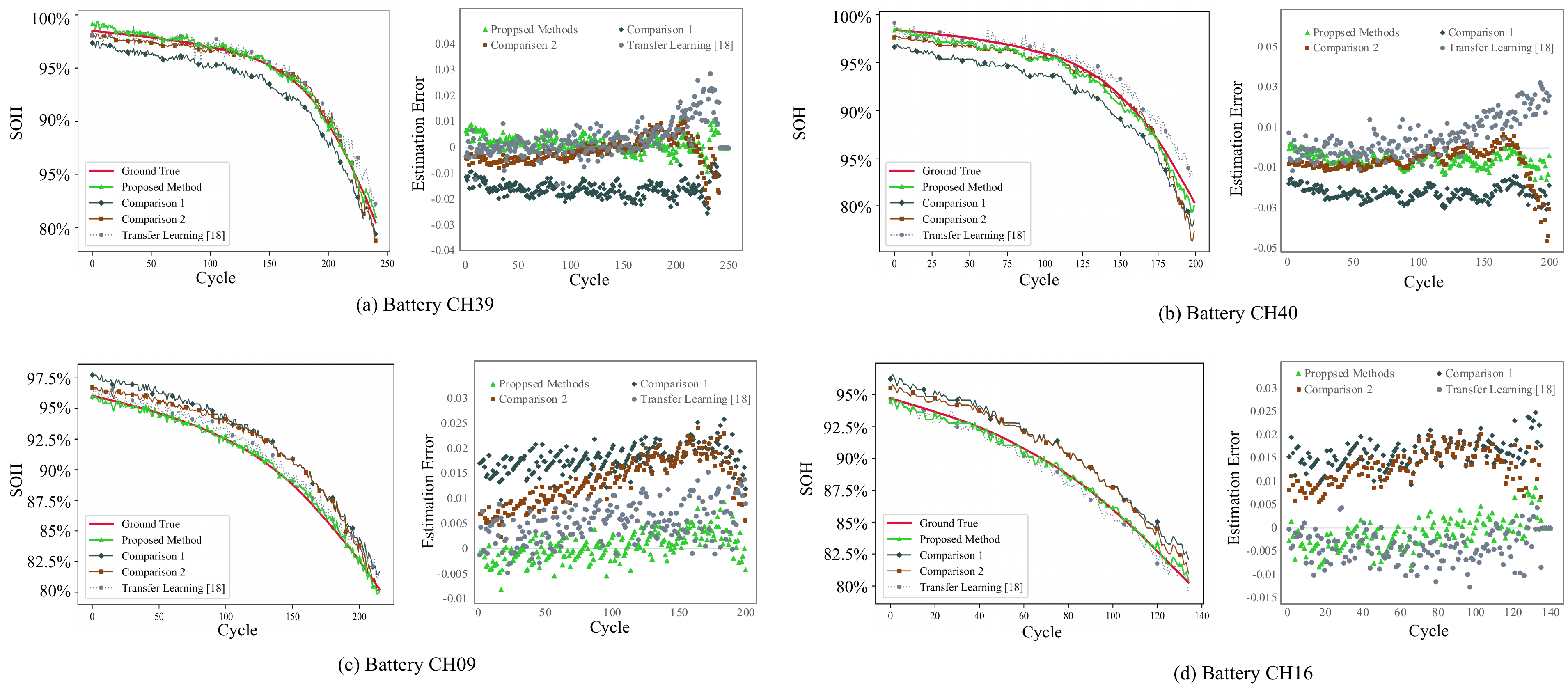}
\caption{Online SOH estimation comparison based on different methods: Based on offline Battery CH17: (a) CH39, (b) CH40; Based on offline Battery CH24: (c) CH09,  and (d) CH16 }
\label{soh}
\end{figure*}

\subsection{Online SOH Estimation}
After training, the models are stored in their respective categories for online SOH estimation. The discharge cycle of the testing battery is monitored following Section III.A.1 and 2. Once the knee onset is detected online, it signals the online battery entering fast degradation.  The online battery is categorized based on the clustering result from the battery pool and chooses the SOH estimation model from the same category.  Table III shows the SOH estimation result range of the proposed methods on the 39 batteries in the battery pool. The proposed method's RMSEs are as low as 0.0022.

To verify the effectiveness of categorization, the models trained in one category are used to estimate the SOH for the online battery from a different category. As observed in Fig. 10(a), the SOH estimation results do not achieve a satisfactory estimation, with the maximum RMSE reaching 0.044. The variance is so large that it is not recommended to use the SOH estimation model from different categories.  In addition, to verify the efficacy of employing data from the $2nd$ stage solely, the models being trained by the entire cycle life, which consists of both the $1st$ stage and the $2nd$ stage data, are used as a benchmark. Using Battery CH24 as an example,  trained by the $2nd$ stage of 161 cycles, the minimum, median, and maximum RMSE are 0.0029, 0.0082, and 0.0220, respectively, compared to the model trained by 690 cycles from the entire cycle life of 0.0048, 0.0187, and 0.0348. As observed in Fig. 10(b) as well, the distribution of prediction errors from the model trained by both stages becomes more pronounced for all the batteries. With much less training data, the proposed method was able to get a better estimation of SOH. 

To further illustrate the effectiveness,  the proposed methods for online state of health (SOH) estimation have been evaluated and compared to existing approaches,  particularly transfer learning. Results obtained from the proposed methods for batteries CH39 and CH40 from the long-range category and batteries CH09 and CH16 from the short-range category, based on offline models trained by batteries CH17 and CH24 respectively, were presented. These results were compared to those using the transfer learning method from literature \cite{48}  as a benchmark.  Fig. 11 plots the estimation results and the estimation error distribution of the online batteries by different methods.  As shown in Table IV,  the proposed methods demonstrated a significant improvement in performance, surpassing not only Comparison 1 and 2, but also outperforming the transfer learning approach.  

\begin{table}[]
\centering
\caption{Comparison between the proposed method and its counterparts regarding the index RMSE}
\begin{tabular}{lllll}\toprule
                             & \multicolumn{2}{c}{Short Range}            & \multicolumn{2}{c}{Long Range}          \\\hline
\\[-1em]
Offline Battery              & CH17                & CH17                & CH24               & CH24                \\\hline
\\[-1em]
Online   Battery             & CH39                & CH40                & CH16               & CH09                \\\toprule
Comparison   1               & 0.0164          & 0.0228          & 0.0165     & 0.0179      \\\hline
\\[-1em]
Comparison   2               & 0.0055          & 0.0107        & 0.0136        & 0.0140        \\\hline
\\[-1em]
Transfer   Learning {[}18{]} & 0.0076        & 0.0112         & 0.0054       & 0.0072        \\\hline
\\[-1em]
Proposed   Method           & \textbf{0.0036} & \textbf{0.0068} & \textbf{0.0035} & \textbf{0.0029} \\\toprule
\end{tabular}
\centering
\end{table}

\section{Conclusion}
The knee onset is an important indicator of battery health as it signifies the commencement of rapid battery degeneration. This research proposes an online knee onset identification method by monitoring the temporal dynamics during battery discharging.  It utilizes dynamic time warping  to extract the temporal information and synchronize the uneven discharge cycles forming the discharge time series. The Matrix Profile identifies the knee onset by performing a similarity search on the discharge time series to reveal the anomaly when the online discharge cycle's nearest neighbor distance surges and the regime changes. Furthermore, the knee onset's contribution to the state of health (SOH) estimation accuracy is validated.  Based on the knee onset and end-of-life cycles distribution, the Gaussian mixture model separates the offline batteries into several categories regarding to the lifespan.  It aids the online battery in selecting the appropriate SOH estimation model from the same category for the improved estimation result without any retraining.  The estimation power of the SOH estimation model is further enhanced by the training data acquired from the fast deterioration period only, which possesses the same statistical distribution.  Our proposed method significantly improves training efficiency and reduces training costs on data collection by requiring less training data than the models trained over the entire cycle life.  In future work, inspired by the successful identification of the online knee onset methodology,  it is worthwhile to explore the detection of the end-of-life cycle for an individual battery using the temporal statistics embedded in the charging and discharging cycles.   In addition, more research is necessary to study the long- and short-range categories that may be used for different datasets to aid in SOH estimation.

\bibliographystyle{IEEEtran}
\bibliography{Reference.bib}

\begin{thebibliography}{10}
\providecommand{\url}[1]{#1}
\csname url@samestyle\endcsname
\providecommand{\newblock}{\relax}
\providecommand{\bibinfo}[2]{#2}
\providecommand{\BIBentrySTDinterwordspacing}{\spaceskip=0pt\relax}
\providecommand{\BIBentryALTinterwordstretchfactor}{4}
\providecommand{\BIBentryALTinterwordspacing}{\spaceskip=\fontdimen2\font plus
\BIBentryALTinterwordstretchfactor\fontdimen3\font minus
  \fontdimen4\font\relax}
\providecommand{\BIBforeignlanguage}[2]{{%
\expandafter\ifx\csname l@#1\endcsname\relax
\typeout{** WARNING: IEEEtran.bst: No hyphenation pattern has been}%
\typeout{** loaded for the language `#1'. Using the pattern for}%
\typeout{** the default language instead.}%
\else
\language=\csname l@#1\endcsname
\fi
#2}}
\providecommand{\BIBdecl}{\relax}
\BIBdecl

\bibitem{52}
X.~Hu, Y.~Che, X.~Lin, and S.~Onori, ``Battery health prediction using
  fusion-based feature selection and machine learning,'' \emph{IEEE
  Transactions on Transportation Electrification}, vol.~7, no.~2, pp. 382--398,
  2021.

\bibitem{57}
A.~Bavand, S.~A. Khajehoddin, M.~Ardakani, and A.~Tabesh, ``Online estimations
  of li-ion battery SOC and SOH applicable to partial charge/discharge,''
  \emph{IEEE Transactions on Transportation Electrification}, vol.~8, no.~3,
  pp. 3673--3685, 2022.

\bibitem{33}
D.~Roman, S.~Saxena, V.~Robu, M.~Pecht, and D.~Flynn, ``Machine learning
  pipeline for battery state-of-health estimation,'' \emph{Nature Machine
  Intelligence}, vol.~3, no.~5, pp. 447--456, 2021.

\bibitem{21}
Y.~Qin, W.~Li, C.~Yuen, W.~Tushar, and T.~Saha, ``IIoT-enabled health
  monitoring for integrated heat pump system using mixture slow feature
  analysis,'' \emph{IEEE Transactions on Industrial Informatics}, vol.~18,
  no.~7, pp. 4725--4736, 2021.

\bibitem{56}
B.~Gou, Y.~Xu, and X.~Feng, ``An ensemble learning-based data-driven method for
  online state-of-health estimation of lithium-ion batteries,'' \emph{IEEE
  Transactions on Transportation Electrification}, vol.~7, no.~2, pp. 422--436,
  2021.

\bibitem{14}
M.-F. Ng, J.~Zhao, Q.~Yan, G.~J. Conduit, and Z.~W. Seh, ``Predicting the state
  of charge and health of batteries using data-driven machine learning,''
  \emph{Nature Machine Intelligence}, vol.~2, pp. 161--170, 2020.

\bibitem{23}
Y.~Qin, C.~Yuen, Y.~Shao, B.~Qin, and X.~Li, ``Slow-varying dynamics-assisted
  temporal capsule network for machinery remaining useful life estimation,''
  \emph{IEEE Transactions on Cybernetics}, pp. 1--15, 2022, doi:
  10.1109/TCYB.2022.3164683.

\bibitem{50}
K.~Liu, X.~Hu, Z.~Wei, Y.~Li, and Y.~Jiang, ``Modified gaussian process
  regression models for cyclic capacity prediction of lithium-ion batteries,''
  \emph{IEEE Transactions on Transportation Electrification}, vol.~5, no.~4,
  pp. 1225--1236, 2019.

\bibitem{20}
Y.~Qin, C.~Yuen, and S.~Adams, ``Invariant learning based multi-stage
  identification for lithium-ion battery performance degradation,'' \emph{The
  46th Annual Conference of the IEEE Industrial Electronics Society}, pp.
  1849--1854, 2020.

\bibitem{27}
R.~Rahimilarki, Z.~Gao, A.~Zhang, and R.~Binns, ``Robust neural network fault
  estimation approach for nonlinear dynamic systems with applications to wind
  turbine systems,'' \emph{IEEE Transactions on Industrial Informatics},
  vol.~15, no.~12, pp. 6302--6312, 2019.

\bibitem{37}
S.~Greenbank and D.~A. Howey, ``Piecewise-linear modelling with automated
  feature selection for li-ion battery end-of-life prognosis,''
  \emph{Mechanical Systems and Signal Processing}, vol. 184, p. 109612, 2023.

\bibitem{12}
X.~Feng, C.~Weng, X.~He, X.~Han, L.~Lu, D.~Ren, and M.~Ouyang, ``Online
  state-of-health estimation for li-ion battery using partial charging segment
  based on support vector machine,'' \emph{IEEE Transactions on Vehicular
  Technology}, vol.~68, no.~9, pp. 8583--8592, 2019.

\bibitem{13}
Z.~Chen, Q.~Xue, R.~Xiao, Y.~Liu, and J.~Shen, ``State of health estimation for
  lithium-ion batteries based on fusion of autoregressive moving average model
  and elman neural network,'' \emph{IEEE Access}, vol.~7, pp.
  102\,662--102\,678, 2019.

\bibitem{11}
Y.~Choi, S.~Ryu, K.~Park, and H.~Kim, ``Machine learning-based lithium-ion
  battery capacity estimation exploiting multi-channel charging profiles,''
  \emph{IEEE Access}, vol.~7, pp. 75\,143--75\,152, 2019.

\bibitem{28}
J.~Zhu, Y.~Wang, Y.~Huang, R.~Bhushan~Gopaluni, Y.~Cao, M.~Heere, M.~J.
  M{\"u}hlbauer, L.~Mereacre, H.~Dai, X.~Liu \emph{et~al.}, ``Data-driven
  capacity estimation of commercial lithium-ion batteries from voltage
  relaxation,'' \emph{Nature communications}, vol.~13, no.~1, pp. 1--10, 2022.

\bibitem{32}
J.~Ma, P.~Shang, X.~Zou, N.~Ma, Y.~Ding, J.~Sun, Y.~Cheng, L.~Tao, C.~Lu,
  Y.~Su, J.~Chong, H.~Jin, and Y.~Lin, ``A hybrid transfer learning scheme for
  remaining useful life prediction and cycle life test optimization of
  different formulation li-ion power batteries,'' \emph{Applied Energy}, vol.
  282, p. 116167, 2021.

\bibitem{22}
Y.~Qin, S.~Adams, and C.~Yuen, ``Transfer learning-based state of charge
  estimation for lithium-ion battery at varying ambient temperatures,''
  \emph{IEEE Transactions on Industrial Informatics}, vol.~17, no.~11, pp.
  7304--7315, 2021.

\bibitem{48}
K.~Q. Zhou, Y.~Qin, and C.~Yuen, ``Transfer-learning-based state-of-health
  estimation for lithium-ion battery with cycle synchronization,''
  \emph{IEEE/ASME Transactions on Mechatronics}, pp. 1--11, 2022, doi:
  10.1109/TMECH.2022.3201010.

\bibitem{7}
X.~Han, M.~Ouyang, L.~Lu, and L.~Jianqiu, ``Cycle life of commercial
  lithium-ion batteries with lithium titanium oxide anodes in electric
  vehicles,'' \emph{Energies}, vol.~7, pp. 4895--4909, 2014.

\bibitem{10}
``IEEE recommended practice for sizing lead-acid batteries for stationary
  applications,'' \emph{IEEE Std 485-2020 (Revision of IEEE Std 485-2010)}, pp.
  1--69, 2020.

\bibitem{34}
K.~Kim, M.~Kim, H.~Churr, G.~Lee, and S.~Han, ``G-K curve-based knee point
  prediction method for li-ion batteries,'' in \emph{2021 21st International
  Conference on Control, Automation and Systems (ICCAS)}, pp. 1190--1193, 2021.

\bibitem{4}
W.~Diao, S.~Saxena, B.~Han, and M.~Pecht, ``Algorithm to determine the knee
  point on capacity fade curves of lithium-ion cells,'' \emph{Energies},
  vol.~12, no.~15, 2019.

\bibitem{5}
S.~Greenbank and D.~Howey, ``Automated feature extraction and selection for
  data-driven models of rapid battery capacity fade and end of life,''
  \emph{IEEE Transactions on Industrial Informatics}, vol.~18, no.~5, pp.
  2965--2973, 2022.

\bibitem{6}
C.~Zhang, Y.~Wang, Y.~Gao, F.~Wang, B.~Mu, and W.~Zhang, ``Accelerated fading
  recognition for lithium-ion batteries with nickel-cobalt-manganese cathode
  using quantile regression method,'' \emph{Applied Energy}, vol. 256, p.
  113841, 2019.

\bibitem{46}
M.~Haris, M.~N. Hasan, and S.~Qin, ``Degradation curve prediction of
  lithium-ion batteries based on knee point detection algorithm and
  convolutional neural network,'' \emph{IEEE Transactions on Instrumentation
  and Measurement}, vol.~71, pp. 1--10, 2022.

\bibitem{2}
P.~Ferm{\'\i}n-Cueto, E.~McTurk, M.~Allerhand, E.~Medina-Lopez, M.~F. Anjos,
  J.~Sylvester, and G.~{dos Reis}, ``Identification and machine learning
  prediction of knee-point and knee-onset in capacity degradation curves of
  lithium-ion cells,'' \emph{Energy and AI}, vol.~1, p. 100006, 2020.

\bibitem{16}
C.-C.~M. Yeh, Y.~Zhu, L.~Ulanova, N.~Begum, Y.~Ding, H.~A. Dau, D.~F. Silva,
  A.~Mueen, and E.~Keogh, ``Matrix profile I: All pairs similarity joins for
  time series: A unifying view that includes motifs, discords and shapelets,''
  in \emph{2016 IEEE 16th International Conference on Data Mining (ICDM)},
 pp. 1317--1322,  2016.

\bibitem{19}
P.~M. Attia, A.~Grover, N.~Jin, K.~A. Severson, T.~M. Markov, Y.-H. Liao, M.~H.
  Chen, B.~Cheong, N.~Perkins, Z.~Yang \emph{et~al.}, ``Closed-loop
  optimization of fast-charging protocols for batteries with machine
  learning,'' \emph{Nature}, vol. 578, no. 7795, pp. 397--402, 2020.

\bibitem{53}
X.~Shu, J.~Shen, G.~Li, Y.~Zhang, Z.~Chen, and Y.~Liu, ``A flexible
  state-of-health prediction scheme for lithium-ion battery packs with long
  short-term memory network and transfer learning,'' \emph{IEEE Transactions on
  Transportation Electrification}, vol.~7, no.~4, pp. 2238--2248, 2021.

\bibitem{49}
Y.~Qin, C.~Yuen, X.~Yin, and B.~Huang, ``A transferable multi-stage model with
  cycling discrepancy learning for lithium-ion battery state of health
  estimation,'' \emph{IEEE Transactions on Industrial Informatics}, 2022,
  doi:10.1109/TII.2022.3205942.

\bibitem{8}
P.~M. Attia, A.~Bills, F.~B. Planella, P.~Dechent, G.~Dos~Reis, M.~Dubarry,
  P.~Gasper, R.~Gilchrist, S.~Greenbank, D.~Howey \emph{et~al.}, ``"Knees" in
  lithium-ion battery aging trajectories,'' \emph{Journal of The
  Electrochemical Society}, vol. 169, no.~6, p. 060517, 2022.

\bibitem{15}
K.~Q. Zhou, Y.~Qin, P.~L.~B. Lau, C.~Yuen, and S.~Adams, ``Lithium-ion battery
  state of health estimation based on cycle synchronization using dynamic time
  warping,'' in \emph{IECON 2021--47th Annual Conference of the IEEE Industrial
  Electronics Society}.\hskip 1em plus 0.5em minus 0.4em\relax IEEE, pp.
  1--6,  2021.

\bibitem{42}
E.~J. Keogh and M.~J. Pazzani, ``Derivative dynamic time warping,'' in
  \emph{Proceedings of the 2001 SIAM International Conference on Data Mining
  (SDM)}, pp. 1--11,  2001.

\bibitem{41}
A.~Kassidas, J.~F. MacGregor, and P.~A. Taylor, ``Synchronization of batch
  trajectories using dynamic time warping,'' \emph{AIChE Journal}, vol.~44,
  no.~4, pp. 864--875, 2004.

\bibitem{18}
Y.~Zhu, S.~Gharghabi, D.~Silva, A.~Dau, C.-C.~M. Yeh, N.~Shakibay~Senobari,
  A.~Almaslukh, K.~Kamgar, Z.~Zimmerman, G.~Funning, A.~Mueen, and E.~Keogh,
  ``The swiss army knife of time series data mining: ten useful things you can
  do with the matrix profile and ten lines of code,'' \emph{Data Mining and
  Knowledge Discovery}, vol.~34, pp. 949---979, 2020.

\bibitem{43}
Y.~Lu, R.~Wu, A.~Mueen, M.~A. Zuluaga, and E.~Keogh, ``Matrix profile XXIV:
  Scaling time series anomaly detection to trillions of datapoints and
  ultra-fast arriving data streams,'' no.~10.\hskip 1em plus 0.5em minus
  0.4em\relax Association for Computing Machinery, pp. 1173--1182, 2022.

\bibitem{47}
F.~Madrid, S.~Imani, R.~Mercer, Z.~Zimmerman, N.~Shakibay, and E.~Keogh,
  ``Matrix profile XX: Finding and visualizing time series motifs of all
  lengths using the matrix profile,'' in \emph{2019 IEEE International
  Conference on Big Knowledge (ICBK)}, pp. 175--182, 2019.

\bibitem{44}
A.~G. Asuero, A.~Sayago, and A.~G. Gonz{\'a}lez, ``The correlation coefficient:
  An overview,'' \emph{Critical Reviews in Analytical Chemistry}, vol.~36,
  no.~1, pp. 41--59, 2006.

\bibitem{26}
Y.~Che, Y.~Zheng, Y.~Wu, X.~Sui, P.~Bharadwaj, D.-I. Stroe, Y.~Yang, X.~Hu, and
  R.~Teodorescu, ``Data efficient health prognostic for batteries based on
  sequential information-driven probabilistic neural network,'' \emph{Applied
  Energy}, vol. 323, p. 119663, 2022.

\bibitem{36}
J.~Yu, ``Health degradation detection and monitoring of lithium-ion battery
  based on adaptive learning method,'' \emph{IEEE Transactions on
  Instrumentation and Measurement}, vol.~63, no.~7, pp. 1709--1721, 2014.

\bibitem{40}
M.~F. Niri, T.~M. Bui, T.~Q. Dinh, E.~Hosseinzadeh, T.~F. Yu, and J.~Marco,
  ``Remaining energy estimation for lithium-ion batteries via gaussian mixture
  and markov models for future load prediction,'' \emph{Journal of Energy
  Storage}, vol.~28, p. 101271, 2020.

\bibitem{24}
S.~Kim, Y.~Y. Choi, K.~J. Kim, and J.-I. Choi, ``Forecasting state-of-health of
  lithium-ion batteries using variational long short-term memory with transfer
  learning,'' \emph{Journal of Energy Storage}, vol.~41, p. 102893, 2021.

\bibitem{25}
Y.~Che, Z.~Deng, X.~Lin, L.~Hu, and X.~Hu, ``Predictive battery health
  management with transfer learning and online model correction,'' \emph{IEEE
  Transactions on Vehicular Technology}, vol.~70, no.~2, pp. 1269--1277, 2021.

\end{thebibliography}

\vspace{12pt}
\end{document}